# Generative Adversarial Networks for Data Generation in Structural Health Monitoring

Furkan Luleci[1], F. Necati Catbas[1]* and Onur Avci[2]

[1]Department of Civil, Environmental, and Construction Engineering, University of Central Florida, Orlando, FL, United States,
[2]Department of Civil, Construction, and Environmental Engineering, Iowa State University, Ames, IA, United States

Structural Health Monitoring (SHM) has been continuously benefiting from the advancements in the field of data science. Various types of Artificial Intelligence (AI) methods have been utilized to assess and evaluate civil structures. In AI, Machine Learning (ML) and Deep Learning (DL) algorithms require plenty of datasets to train; particularly, the more data DL models are trained with, the better output it yields. Yet, in SHM applications, collecting data from civil structures through sensors is expensive and obtaining useful data (damage associated data) is challenging. In this paper, one-dimensional (1-D) Wasserstein loss Deep Convolutional Generative Adversarial Networks using Gradient Penalty (1-D WDCGAN-GP) is utilized to generate damage-associated vibration datasets that are similar to the input. For the purpose of vibration-based damage diagnostics, a 1-D Deep Convolutional Neural Network (1-D DCNN) is built, trained, and tested on both real and generated datasets. The classification results from the 1-D DCNN on both datasets resulted in being very similar to each other. The presented work in this paper shows that, for the cases of insufficient data in DL or ML-based damage diagnostics, 1-D WDCGAN-GP can successfully generate data for the model to be trained on.

## INTRODUCTION

During the operational life of civil structures, different types of damages can shorten the remaining useful life of the structures. This is particularly important in today's world, where catastrophic events are increasing, and they are projected to be frequent in the future. In addition, most civil structures that were built decades ago have already started losing their functionality and capacity. This is especially true for the US, where recent studies show the substandard conditions of existing civil structures. Therefore, it is essential to diagnose the structural damages and subsequently prognose the remaining useful life and, based on that, implement an effective health management plan to improve the life cycle of the structure. Hence, with correct maintenance action items derived from the health management plan, the structural lifetime can be extended, costly structural failures can be avoided, and, more importantly, human lives can be saved.

## Brief Review on Structural Damage Diagnostics

In the field of Structural Health Monitoring (SHM), the common practice to assess an existing civil structure is to collect operational data using sensors such as accelerometers, potentiometers, strain gauges, fiber optic sensors, or load cells. Damage identification is performed by monitoring the changes in structural properties like stiffness, mass, and damping in an attempt to detect and locate structural defects such as cracks, delamination, spalling, corrosion, and bolt loosening. In vibration-based applications, acceleration data are predominantly used because it has advantages such as being easy to use and being efficient in data processing (Catbas and Aktan, 2002; Catbas et al., 2006; Das et al., 2016).

Vibration-based structural damage diagnostics can be executed in two ways: 1) local methods where Non-Destructive Testing (NDT) and some camera sensing techniques (IR, DIC, RGB, etc.) are involved, and 2) vibration-based (global) methods where the collected vibration data are analyzed parametrically (using a physical model like FEA software or non-physical model like system identification algorithms to estimate the structure's physical parameters) or nonparametrically (using statistical approaches on the raw vibration data to find the features and then classify or detect the anomalies in it) (Catbas and Malekzadeh, 2016; Avci et al., 2021). In addition, using advanced computer vision techniques, SHM can be conducted at local and global levels, and then, damage diagnostics can be performed (Dong et al., 2021).

With the goal of diagnosing structural damages in the civil structures, several studies are presented in the SHM field by Krishnan Nair and Kiremidjian (2007), Gul and Catbas (2008), Yin et al. (2009), Gul and Catbas (2011), and Silva et al. (2016). The abovementioned studies are parametrically and nonparametrically conducted vibration-based structural damage diagnostic works. One of the common details in those studies is that the amount of collected data is not influential on the success of the proposed algorithms, unlike the Machine Learning (ML) and Deep Learning (DL) algorithms where they require plenty of data input—particularly, DL algorithms yield exceptionally sound results on as much data as possible (Alom et al., 2019).

With the emergence of ML and DL algorithms, they have been used in SHM due to their high performance on feature extraction, classification, regression, and clustering techniques. Several ML methods including parametric and nonparametric vibration-based damage diagnostic studies are introduced. The Artificial Neural Network (ANN) is observed to be the most used algorithm in an ensemble or integrated with different algorithms, followed by Support Vector Machine (SVM). Some of them are those in the works of Lee et al. (2005), Lee and Kim (2007), González and Zapico (2008), Cury and Crémona (2012), Gul et al. (2014), Bandara et al. (2014), Abdeljaber and Avci (2016), Ghiasi et al. (2016), and Santos et al.(2016). Furthermore, it is essential to note that, apart from parametric-based methods where they rely on system identification techniques to extract the structural parameters, using an ML model for nonparametric-based damage diagnostic requires the usage of feature extraction from the raw data (Catbas and Malekzadeh, 2016) such as Principal Component Analysis or Autoregressive. Thus, this causes computational complexity and time, along with other limitations (Avci et al., 2021). The DL algorithms can learn to extract useful features from the raw data and train on them to make accurate predictions. In short, with a correctly built model and proper training, DL models can show superior performance over ML models. In the civil SHM field, there are few studies, including vibration-based unsupervised DL—mostly Autoencoders (Pathirage et al., 2018; Shang et al., 2021; Rastin et al., 2021), and few supervised DL—mostly Convolutional Neural Networks (Abdeljaber et al., 2017, 2018; Eren, 2017; Avci et al., 2017; Yu et al., 2019).

## Motivation and Objective

In civil SHM, data collection is a challenging and expensive task such as getting permission from authorities to install costly and laborious SHM systems, setting up communication networks between sensors and data acquisition systems, requesting traffic closures, and having skilled experts on the field. In addition, obtaining valuable data containing damage-associated features is not always very easy. This is one of the previously mentioned SHM challenges. Considering that only a few civil structures have permanent SHM systems in the world, it is also challenging to know about the damage state of the remaining structures. As such, data scarcity is a challenge in civil SHM. One solution is oversampling the dataset by increasing the copy of the existing dataset. However, this solution does not teach the Artificial Intelligence (AI) models to learn the variation in the damage-associated data but teaches only the provided one to the model and may lead to overfitting. Building an FEA model and analyzing the structure under similar damage scenarios then producing displacement, stress, or acceleration data are another solution to tackle the data scarcity challenge. Yet, this methodology can be inaccurate and possibly unreliable compared to real data from particularly complex structures. In addition, it is implausible that observed damages can be reflected correctly in a model along with accumulated numerical FEA errors (Gardner and Barthorpe, 2019).

As the focus shifted toward ML and DL models for the damage diagnostics of civil structures, the aforementioned data scarcity problem hinders the usage of these algorithms because they require large datasets. The study presented in this paper employs Generative Adversarial Networks (GANs) to generate valuable data to be further used by a Deep Convolutional Neural Network (DCNN) model to perform nonparametric damage diagnostics on the existing data of a steel laboratory structure. Specifically, the presented study investigates vibration-based damage detection with scarce data, where GANs can generate data for the ML or DL model to be trained on and then perform damage identification. The proposed methodology in this paper will pave the way for more AI-based SHM procedures to be performed when the data are scarce.

## Background on the GAN

Goodfellow et al. (2014) introduced a novel framework containing two separate networks: (i) a generative model that captures the given random data distribution, $z$, and tries to maximize the likeliness of the output it produces, $G_\theta(z)$, very similar to the training input, $x$, and (ii) a discriminative model, $D_\varphi$, that tries to maximize the possibility of the data it receives from the generator, $G_\theta(z)$, is fake and from the training input, $x$, is real. In that manner, it is considered a two-player game where each side is trying to deceive the other. The loss formulation to train the GAN in the paper is shown in **Eq. 1**.

$$\min_\theta \max_\varphi V(G_\theta, D_\varphi) = E_{x \sim pdata(x)} \left[ log D_\varphi(x) \right]$$
$$+ E_{z \sim pz(z)} \left[ log\left( 1 - D_\varphi(G_\theta(z)) \right) \right] \quad (1)$$

The methodology is used successfully on image-based applications. Yet, training GANs might be the most challenging among the other DL networks. For instance, they are very hard to converge due to finding a unique solution to Nash equilibrium where the optimization process is looking to find a balance between two sides instead of a minimum. GANs often experience large oscillation in loss values of generator and discriminator during the training, making it harder to reach convergence. Another issue is the mode collapse where the generator part produces the same outputs due to learning a feature in the data that can be used to trick the discriminator easily. In addition, intuitively, GANs suffer from the discriminator being powerful over the generator, which causes the generator training to fail due to vanishing gradients; thus, the discriminator does not provide sufficient information for the generator to learn (Goodfellow, 2017; Salimans et al., 2016). There are several "hacks" introduced to alleviate these drawbacks in the DL community and some of them are used in this study as well as discussed in the following sections. Radford et al. (2016) proposed using the DCNN in GANs after their adoption in computer vision applications. In their study, they noticed that, in the training process, DCNN helped the GAN to learn significantly. Yet, Arjovsky et al. (2017) introduced a GAN that uses Wasserstein distance as a loss function (WGAN), which improves the training of GAN. In their network, instead of using a discriminator that estimates the probability of the generated images as being real or fake, they used a critic which scores the output's realness or fakeness of a given image. Fundamentally, WGAN seeks a minimization of the distance between the generated and the training data distribution. WGAN showed considerable benefits to training the GAN, such as being more stable and less sensitive to the parameters and model architecture; loss functions are more meaningful as they directly relate to the quality of generated images. Gulrajani et al. (2017) proposed using a penalization of the gradient during the training of the critic due to using weight clipping on the critic, which enforces the Lipschitz constraint and therefore lowers the learning capacity of the model. They named the model Wasserstein Generative Adversarial Networks with Gradient Penalty (WGAN-GP). The authors showed that the proposed method performed better than WGAN and provided more stable training.

GANs are primarily used in the computer vision field which processes two-dimensional (2-D) data. In addition, there are some studies in different disciplines that attempted to use GANs for one-dimensional (1-D) data generation and reconstruction for different purposes (Truong and Yanushkevich, 2019; Kuo et al., 2020; Luo et al., 2020; Wulan et al., 2020; Sabir et al., 2021; Wang et al., 2021). In the SHM field of non-civil structures, some studies of GAN-based 1-D data generation, reconstruction, and then training an ML classifier are introduced (Gao et al., 2019; Shao et al., 2019; Guo et al., 2020; Zhang et al., 2021). Few studies are introduced in the SHM field of civil structures related to using GAN for 1-D data reconstruction (Zhang et al., 2018; Fan et al., 2021; Jiang et al., 2021). However, no study investigated the usage of WGAN and WGAN-GP extensively to address the data scarcity problem for civil structures and test the generated synthetic data samples on a DCNN model for damage diagnostics on the raw vibration data. The provided methodology will enable the researchers and professionals to generate the needed data samples to train the DL model to be used for vibration-based damage diagnostics.

## METHODOLOGY

The authors of this study used WGAN-GP that is built on a DCNN, which is named WDCGAN-GP. Because the data used in this study is 1-D, all the convolution operations are executed 1-D both for WDCGAN-GP and DCNN. Thus, in short, 1-D WDCGAN-GP is utilized to generate a synthetic dataset and 1-D DCNN to perform nonparametric damage identification. For simplicity, in the rest of the paper, 1-D WDCGAN-GP and 1-D DCNN is referred to as $\mathcal{M}_1$ and $\mathcal{M}_2$, respectively. The workflow followed in this study can be summarized in this order: 1) data preprocessing for $\mathcal{M}_1$, 2) building the $\mathcal{M}_1$, 3) training and fine-tuning the $\mathcal{M}_1$, 4) evaluation and interpretation of results from the $\mathcal{M}_1$, 5) data preprocessing for the $\mathcal{M}_2$, 6) building the $\mathcal{M}_2$, 7) training and fine-tuning the $\mathcal{M}_2$, 8) testing the $\mathcal{M}_2$, and 9) evaluation and interpretation of the results from the $\mathcal{M}_2$. The dataset, technical notations, and workflow are explained in the following paragraphs.

The vibration dataset is obtained from the study conducted by Abdeljaber et al. (2017) on a steel laboratory frame (**Figure 1**) where a total of 30 accelerometers were installed at each 30 joints and a modal shaker excitation is applied on the structure. Respectively, 30 different damaged and one undamaged scenarios are created at 30 joints separately by loosening the bolts of the connections between filler beams and girders. Then, they collected 256 s of vibration data at a sampling rate of 1,024 Hz with a total sample of 262,144 for each damaged case (a total of 30 cases). The tensor notation used in this paper is $n[a_{xyf}]_s$, where $x$ represents the condition such as 0 means the data are collected in an undamaged scenario and 1 means the data are damaged scenario, $y$ represents the joint number where the vibration data are collected, and $f$ refers to "fake" if it is generated by the $\mathcal{M}_1$. The $s$ refers to the number of samples that a tensor contains, and $n$ refers to "number" of tensors; if there is no number, then it is one tensor. This study worked on only two different sizes of sample. To represent the size of tensors in a simple way, 262,144 and 1,024 samples are denoted as $r$ and $b$, respectively, where $r$ refers to "raw" and $b$ refers to "batch". The generated data from the $\mathcal{M}_1$ are a sample size of 262,144; it is also the used batch size in the model. For DCNN, to use the raw and generated tensors together, raw tensors are also divided into 1,024 size batches randomly (note that the output tensors from the GAN are also generated randomly because the input, $[a_{11}]_r$, is batch sampled in the shuffle during the training). A similar method is also successfully implemented in a study by Abdeljaber et al. (2017), where they divided the raw vibration signal into frames and shuffled it randomly. Besides, because the purpose of this study is to demonstrate that GANs can tackle the data scarcity problem for nonparametric damage diagnostics for civil structures, the sequence or the parameters of the signal tensors are irrelevant. The PC used in this study has the following specs: 16 GB RAM DD4 2933 MHz and NVIDIA GeForce RTX 3070 8 GB GDDR6 graphic card.

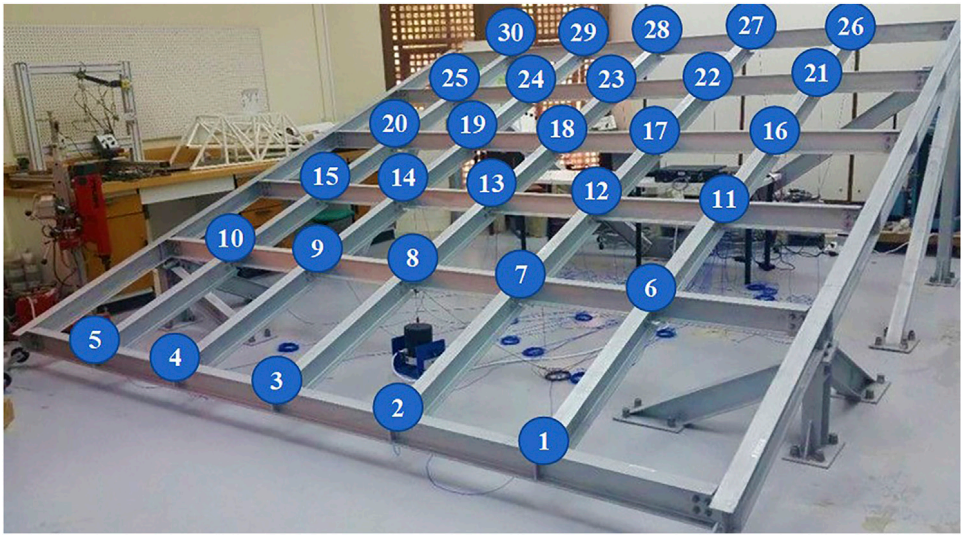

**FIGURE 1 |** Steel frame grand simulator structure (Abdeljaber et al., 2017).

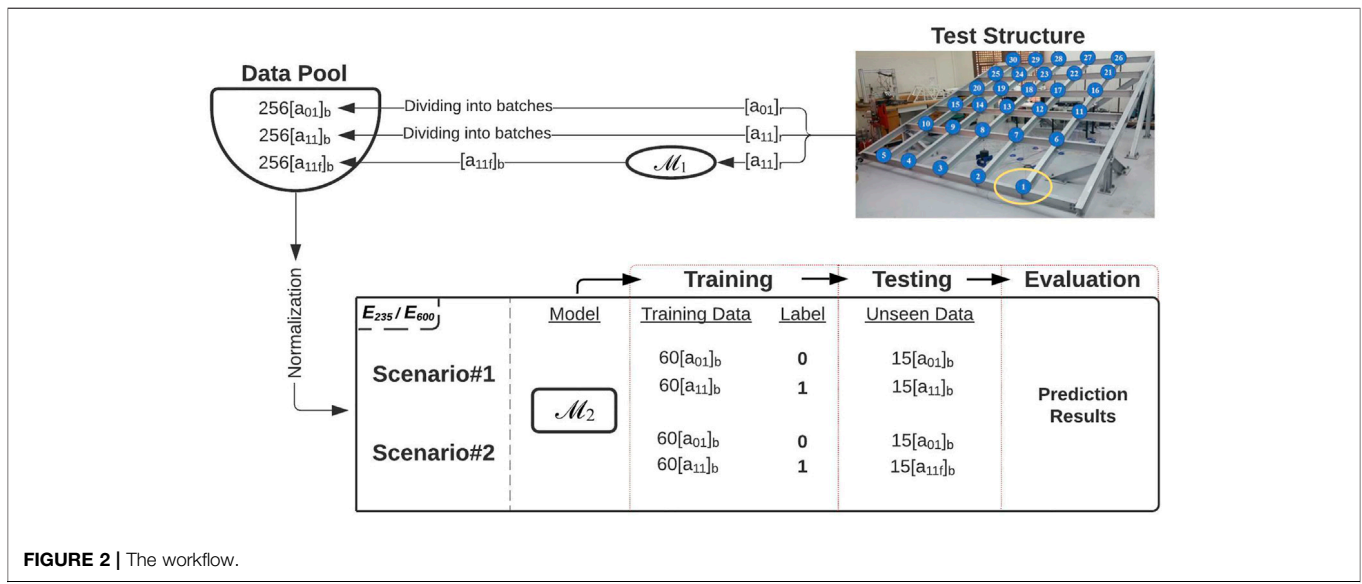

**FIGURE 2 |** The workflow.

**Figure 2** represents the workflow of this study where tensors $[a_{01}]_r$ and $[a_{11}]_r$ are divided into 256 batches each. Then, $[a_{01}]_b$ and $[a_{11}]_b$ are extracted to the data pool. The tensor $[a_{11}]_b$ inputted in the $\mathcal{M}_1$ model to generate synthetic data of $256[a_{11f}]_b$. Consequently, these tensors are used in $\mathcal{M}_2$ for two different scenarios. In other words, $\mathcal{M}_1$ generates the synthetic data samples, $[a_{11f}]_b$, to be tested by $\mathcal{M}_2$ along with the real undamaged samples, $[a_{01}]_b$. The test dataset is taken as 25% of the training data in each scenario. In Scenario#1, normal conditions are assumed in which the undamaged and damaged datasets are real tensors. To represent that, $60[a_{01}]_b$ and $60[a_{11}]_b$ are used in training and $15[a_{01}]_b$ and $15[a_{11}]_b$ are used in the testing dataset. In Scenario#2, for training the model, real undamaged and damaged tensors are used, but for testing, generated damaged along with real undamaged tensors are used. To represent that, $60[a_{01}]_b$ and $60[a_{11}]_b$ are used in training and $15[a_{01}]_b$ and $15[a_{11f}]_b$ are used in testing. Scenario#1 is created to form a reference line for comparison with the other scenarios. Scenario#2 aims to demonstrate the performance of $\mathcal{M}_1$ as to how accurate can $\mathcal{M}_2$ be trained on real data and make correct predictions on generated and real data. In other words, it is to evaluate to what extent $\mathcal{M}_2$, which was trained using real data, can make correct predictions on the fake damaged and real undamaged data. Furthermore, the workflow shown in **Figure 2** is repeated on the same $\mathcal{M}_1$ configuration twice for validation and to investigate the effect of more training on the outputs as well as on the decision by $\mathcal{M}_2$. Hence, the case where $\mathcal{M}_1$ is trained with fewer epochs (iterations) is denoted as $E_{235}$ and the one with more epochs as $E_{600}$. The next sections explain the $\mathcal{M}_1$ and $\mathcal{M}_2$ workflow in detail.

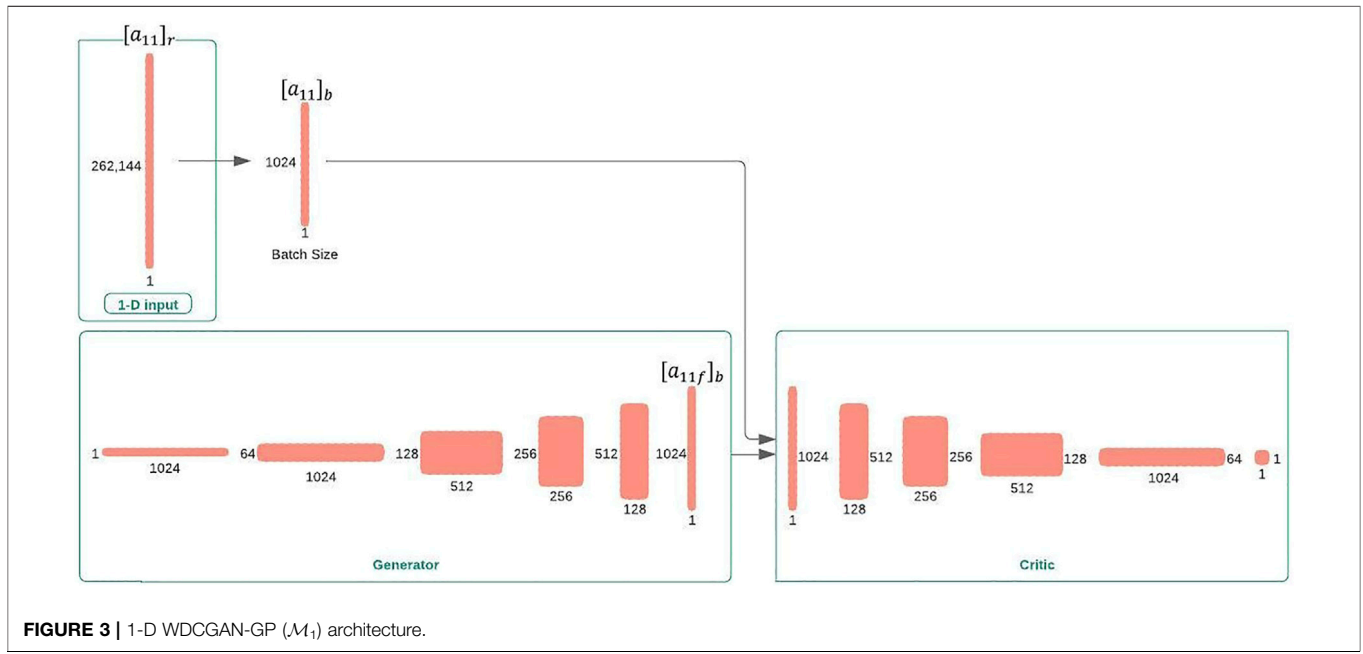

**FIGURE 3 |** 1-D WDCGAN-GP ($\mathcal{M}_1$) architecture.

## $\mathcal{M}_1$–Data Preprocessing

The common practice for DL models before training is to make datasets in the same scale for the model to learn and predict more efficiently. In addition, during the front and back propagation through the network where the dot product of weights is calculated, normalization would help the model by putting the datasets on the same scale to alleviate the impact of spike values; thus, it helps the model to yield more accurate results and less computation time. Otherwise, the spikes would have a huge impact on the weight propagation that can result in extreme values, which lowers the model quality in training. In addition, the $\mathcal{M}_1$ model consists of batch and instance normalization layers that normalize the data batches during the training. Considering that the dataset does not contain large spikes, the model is tried with both normalized and raw training input, and it is observed that results are not distinctly different. In fact, it is believed that the model captures the spatial-temporal features better when it receives the raw vibration dataset. Hence, normalization is not implemented before the training.

## $\mathcal{M}_1$–Architecture

After several trials of different model architectures, the one used in this study is shown in **Figure 3**. First, the generator receives the dimensional noise tensor ($z$) and passes it through five 1-D transpose convolutions, in which the first layer is $filter = 64$, $stride = 2$, $padding = 0$, and then, the rest are $filter = 4$, $stride = 2$, $padding = 1$. Batch Normalization and then ReLU are used after every convolution layer except the last one. After the last convolution layer, Tanh function is used. Consequently, $[a_{11f}]_b$ is created. Following this, the training input $[a_{11}]_b$ and the generated tensor $[a_{11f}]_b$ are passed to the critic (which is called the "discriminator" in GAN but in WGAN, it is "critic") to be evaluated as to how real or fake each tensor is. The critic takes the tensor and passes it through five 1-D convolutions, in which the first four layers are $filter = 4$, $stride = 2$, $padding = 1$ and the last layer is $filter = 64$, $stride = 2$, $padding = 0$. After the first convolution layer, Leaky ReLU and then Dropout is used. Then, after every second, third, and fourth convolution layers, Instance Normalization and then Leaky ReLU are used.

## $\mathcal{M}_1$–Training and Fine-Tuning

As mentioned, GANs are arguably the most difficult DL models to train among other DL models. Therefore, it requires considerable effort to fine-tune. Although the model used in this study, 1-D WDCGAN-GP, is one of the most robust GAN models in the literature, few approaches have been taken during the fine-tuning after many trials with different hyperparameters.

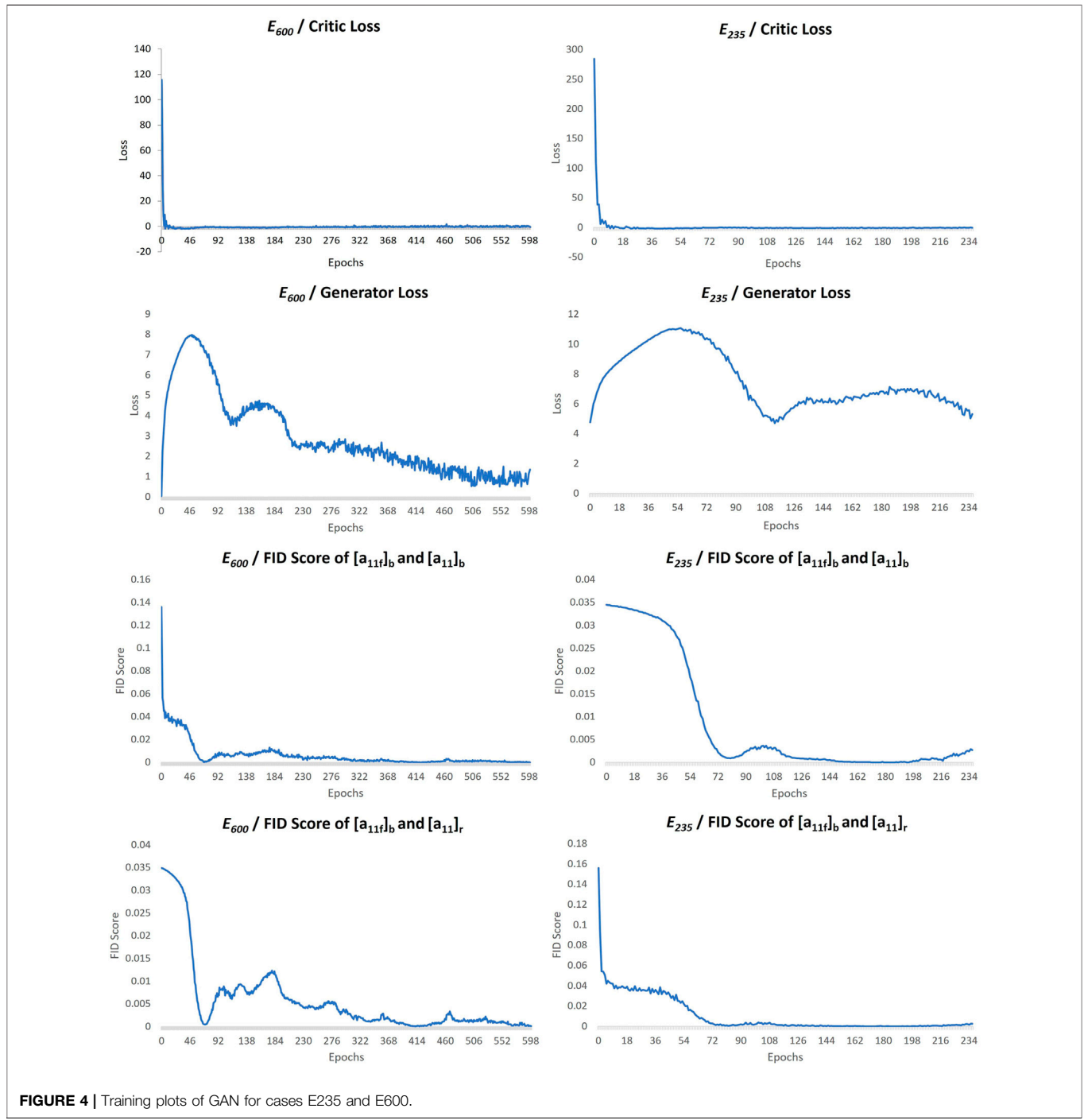

**FIGURE 4 |** Training plots of GAN for cases E235 and E600.

First, using one layer of dropout with 70% in the critic was found beneficial in training. Thus, the capacity of the critic reduces and balances the equilibrium between two adversarial networks and avoids the overfitting problem. Next, a decaying random Gaussian noise is added to the training input to decrease the learning rate of the critic. The learning rate of $5 \times 10^{-6}$ for generator and $2 \times 10^{-5}$ for critic gave the best result. The critic iterations, lambda parameter for the gradient penalty, and batch size are picked as 12, 20, and 1,024, respectively. The number of epoch used is 235 (case $E_{235}$) and 600 for (case $E_{600}$). It took 18 and 44 h of training, respectively. Last, AdamW optimizer is used in both the generator and critic.

## $\mathcal{M}_1$–Evaluation and Interpretation of Results

Evaluation of the GAN models can be categorized as qualitative and quantitative evaluation, where the former is based on visual

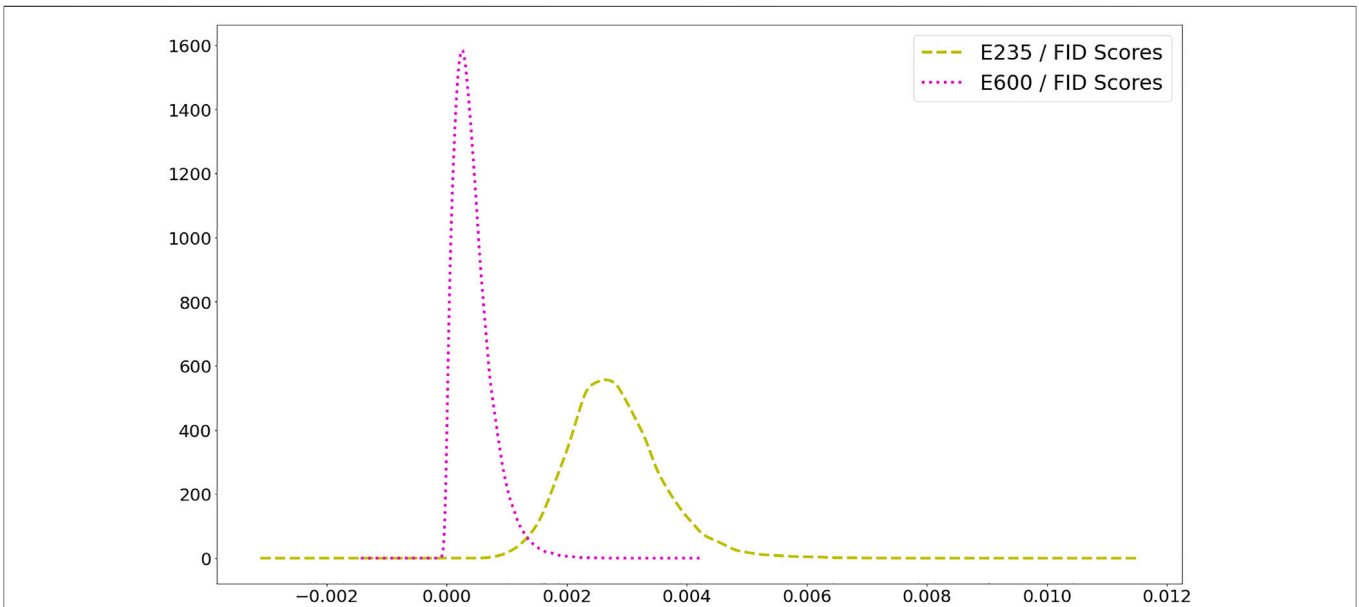

**FIGURE 5 |** The plot of Probability Density Functions of FID scores for $E_{235}$ and $E_{600}$.

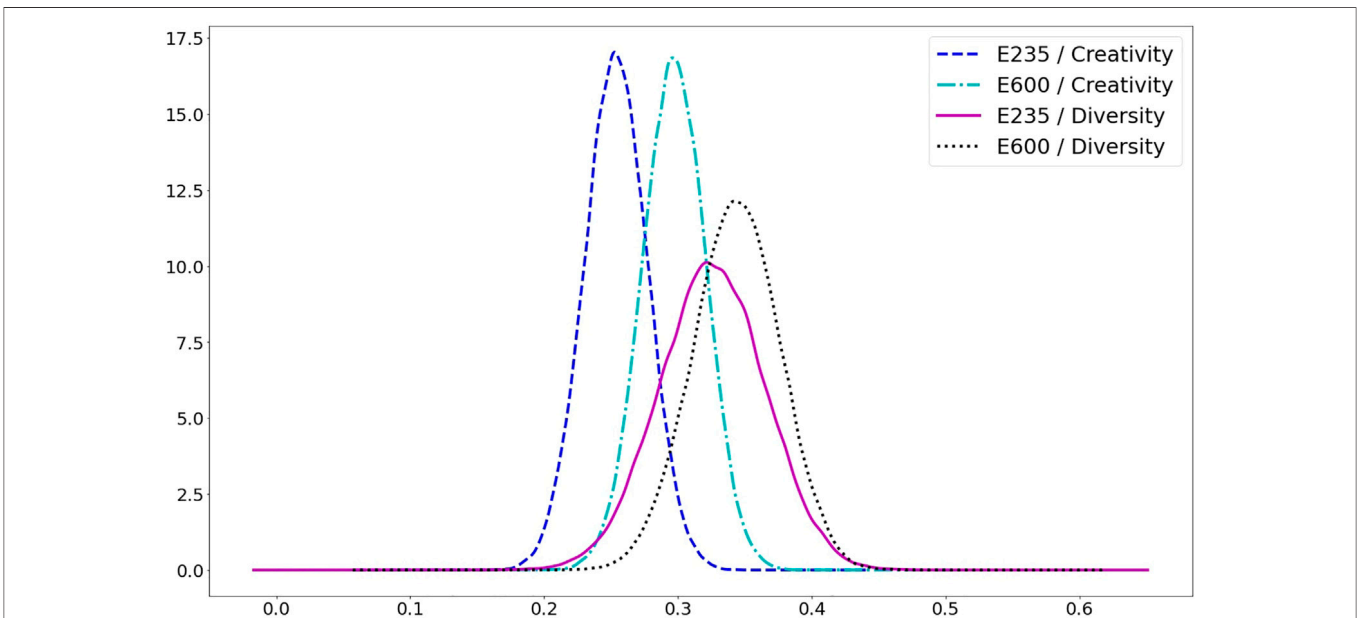

**FIGURE 6 |** The plot of Probability Density Functions of Creativity and Diversity measures for cases $E_{235}$ and $E_{600}$.

evaluation, and the latter is based on numerical evaluation. The most used form of evaluation of GANs used in the DL field is visually comparing the generator's output with the training data, which are primarily images. Yet, this qualitative approach might not be an easy or efficient way for 1-D data as it suffers from some limitations, such as a limited number of generated output can be viewed by an observer in limited time or observed subjectively by different observers. In addition, unlike the other DL models, GANs lack objective function that makes it challenging to evaluate the performance of the model. That is why there are several methods to assess the model's performance with no consensus in the DL field yet as to which quantitative measure is the most effective. The study from Borji (2018) investigated the evaluation methods of GANs, and readers are directed to that reference. Recently, the Fréchet Inception Distance (FID) score has been introduced (Heusel et al., 2017) and has become the most used quantitative evaluation method for GANs as several studies proved its effectiveness against other methods such as

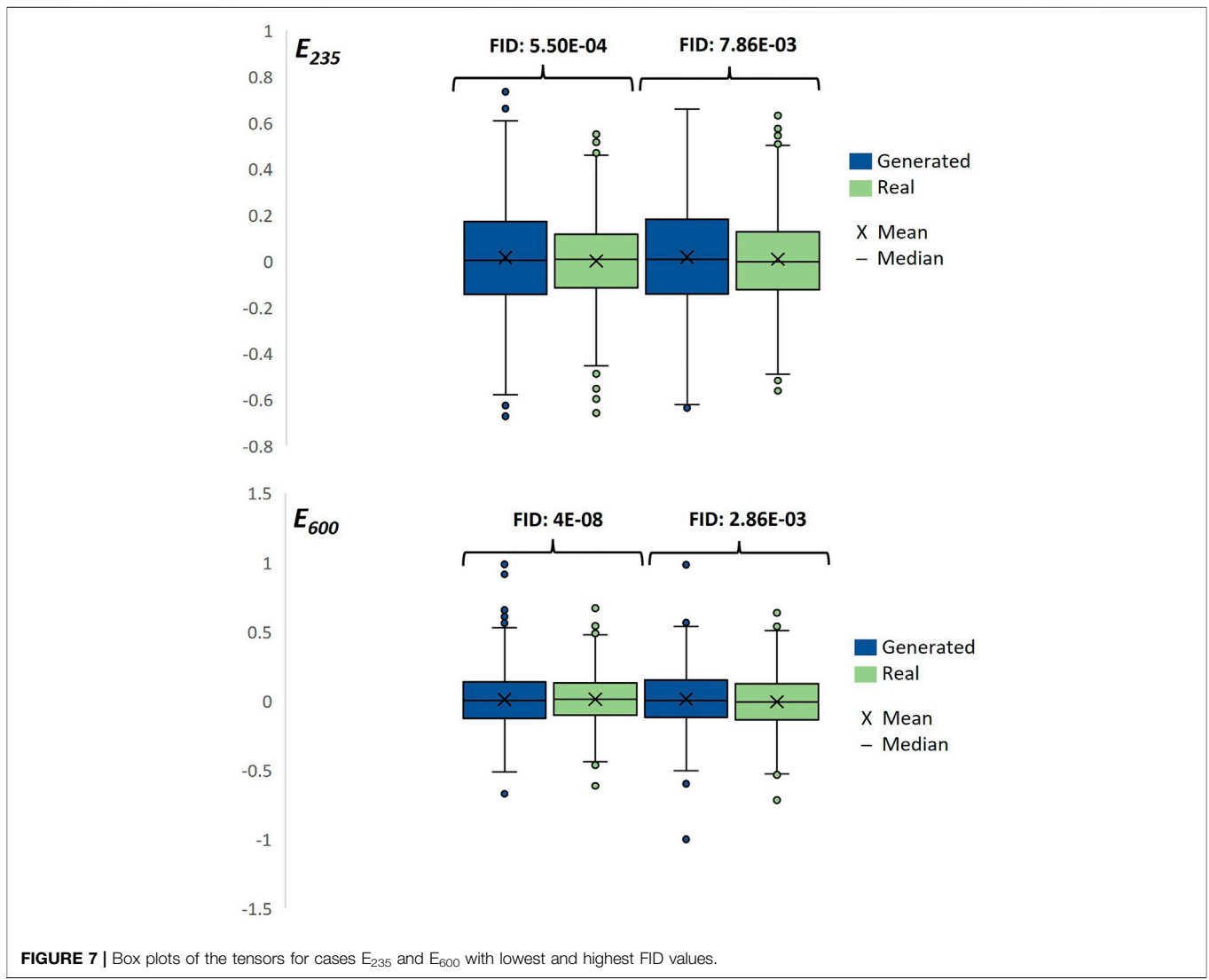

**FIGURE 7 |** Box plots of the tensors for cases $E_{235}$ and $E_{600}$ with lowest and highest FID values.

Inception Score (IS). The FID score is introduced as an improvement over the IS, which lacks capturing the similarity of real input to the produced output. Particularly, the FID has shown remarkably consistent results when compared with a qualitative evaluation of the GAN outputs. The FID formula is based on a statistical formulation, which is provided in **Eq. 1**.

$$FID(x, g) = \left\| \mu_x - \mu_g \right\|_2^2 + Tr\left( C_x + C_g - 2\left( C_x C_g \right)^{0.5} \right) \quad (2)$$

where $\mu_x$ and $\mu_g$ are the means; $C_x$ and $C_g$ are the covariance matrices of real and generated signals, respectively; and $Tr$ is the trace of the matrices, e.g., the sum of all the diagonal elements in the matrices. The lower the FID score, the more similar the compared datasets. Structural Similarity Index Measure (SSIM) (Wang et al., 2004) is another quantitative evaluation metric that is mainly used for the image quality assessment of two images based on a statistical formulation. For instance, if the images are the same, then the SSIM is 1, and if they are different, then it is 0. A paper by Sabir et al. (2021) conducted a successful study on signal data where they evaluated the GAN outputs on the basis of the creativity and diversity approaches using SSIM. For example, creativity indicates that generated signals are not the exact ones of the real data, and diversity demonstrates how the generated datasets look like each other. In their study, they used the threshold of 0.8 for investigating creativity, and if the SSIM of two signals is higher than 0.8, then it is concluded that signals are duplicates. These measures are essential for evaluating the GANs as they are expected to add creativity and diversity to the outputs (Guan and Loew, 2020). Yet, there is no consensus on what SSIM value is good for creativity and diversity for evaluating GANs. In this study, the creativity approach is investigated considering the threshold value of 0.8; in addition, the diversity approach is explored on the basis of how the SSIM values are close to 1 and 0. Whereas the creativity is explored by using the SSIM of the generated and real tensors and the diversity is explored by using the SSIM of the generated tensors within the entire generated dataset. The SSIM formula is given in **Eq. 2**.

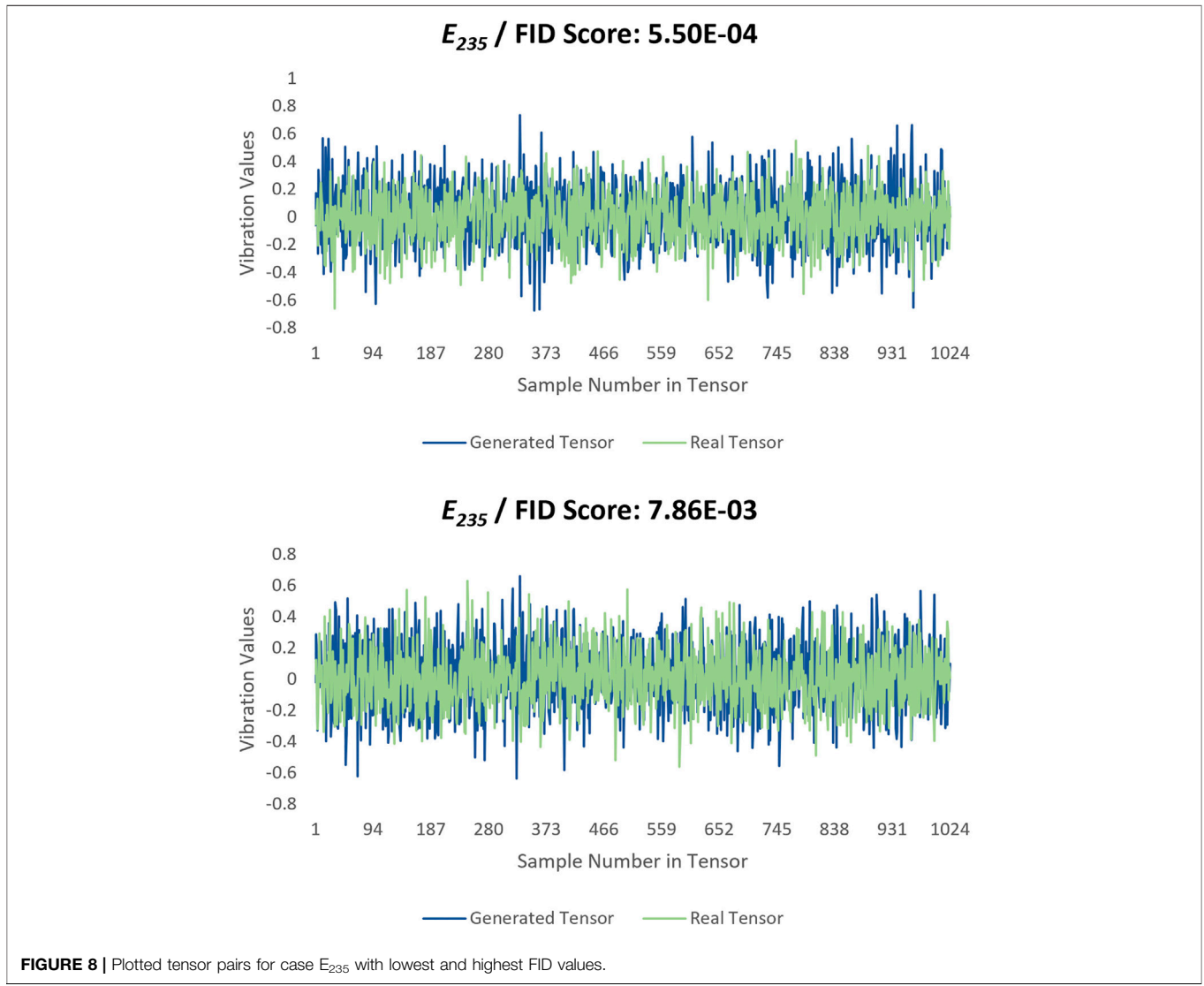

**FIGURE 8 |** Plotted tensor pairs for case $E_{235}$ with lowest and highest FID values.

$$\text{SSIM}(x, g) = \frac{(2\mu_x \mu_g + c_1)(2\sigma_{xg} + c_2)}{(\mu_x^2 + \mu_g^2 + c_1)(\sigma_x^2 + \sigma_g^2 + c_2)} \quad (3)$$

where $\mu_x$ and $\mu_g$ are the means; $\sigma_x$ and $\sigma_g$ are the standard deviations; and $\sigma_{xg}$ is the covariance of real data ($x$) and generated data ($g$), respectively. The $c_1$ and $c_2$ are the constants that are multiplication of $k_1$ and L and of $k_2$ and L, respectively, to stabilize the division with a weak denominator where L is the dynamic range of the signal and $k_1$ and $k_2$ are the constants that are chosen in this study as 0.01 and 0.03, respectively.

During the training of both cases, $E_{235}$ and $E_{600}$, the critic and generator losses and FID scores are plotted to monitor the learning process of the model. The critic losses are converging near zero. The generator losses are first increasing as the one that is superior, the critic, rejecting outputs of the generator because the critic has more knowledge on the real data domain. However, after the generator starts learning the gradients and producing more real-looking datasets, it is seen to be returning to its first loss value, which is expected for WGAN-GPs (**Figure 4**, Generator Loss–tagged plots). The FID calculations are made between the batches, $[a_{11f}]_b$ and $[a_{11}]_b$, and they seem to be reduced to zero for both cases, which means they are getting similar to each other (**Figure 4**, FID-tagged plots).

To ensure that $[a_{11f}]_b$ is not only getting similar to $[a_{11}]_b$ but also to $[a_{11}]_r$, which is the input of $\mathcal{M}_1$, the FID between $[a_{11f}]_b$ and $[a_{11}]_r$ is computed. The training results show that the FID scores between $[a_{11f}]_b$ and $[a_{11}]_b$, and the FID scores between $[a_{11f}]_b$ and $[a_{11}]_r$ are following a similar path during the training (**Figure 4**, FID-tagged plots). This also reveals that the sampled batches, $[a_{11}]_b$, from the real input, $[a_{11}]_r$, have repetitive features in a particular sample length in 262,144 samples. Thus, it validates the study to continue using the 1-s tensors such as $[a_{11f}]_b$, $[a_{11}]_b$, and $[a_{01}]_b$. Because the aim is nonparametric damage diagnostics, it is not based on any parameters but on raw data; hence, the order of the samples in datasets is irrelevant. Therefore, the method of sampling with 1-s batches in shuffle mode from the original input is used, which helps the training of the model for faster convergence, preventing bias, preventing learning the order of the data, etc. Thus, for the rest of the study, only the calculations between batch sampled generated and real data ($[a_{11f}]_b$ and $[a_{11}]_b$) are considered. Furthermore, during the training, in case $E_{235}$, the FID score started from 0.0345 and decreased to 0.0021, which means a reduction of around 16 times, and in case $E_{600}$, the FID score started from 0.1359 and decreased to 0.000015, a reduction of 9,060 times. The large difference in the reduction value shows that the number of training epochs helped the model to learn the dataset thoroughly. For a better understanding of the FID score values, a study by Costa et al. (2019) can be informative. Costa et al. compared their GAN model to others on the MNIST dataset and the FID score decreased by six times during the training. Although the 9,060 times decrement can be viewed as a success, it can also be an indication of overfitting for case $E_{600}$, which will be investigated in the following paragraph.

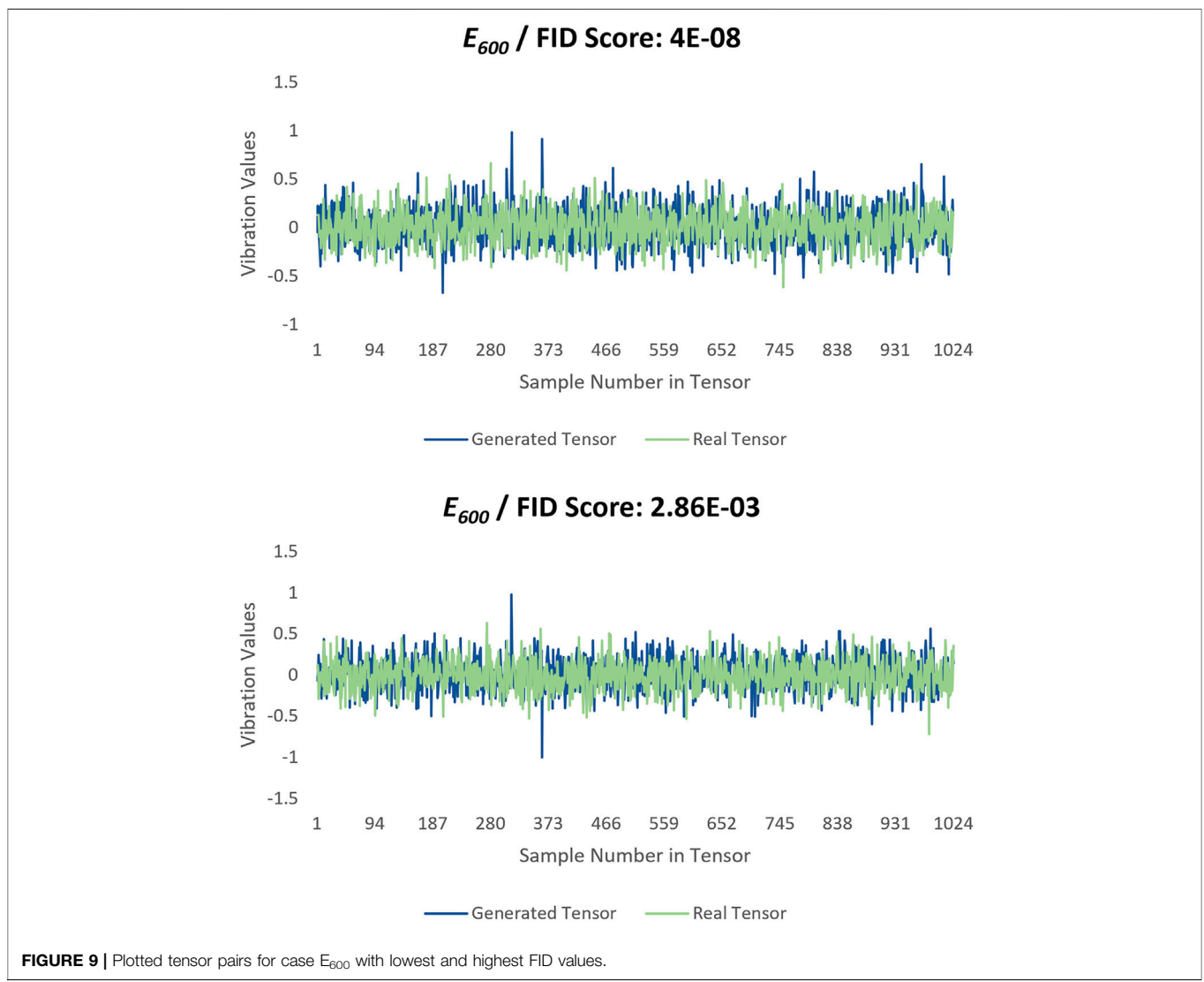

**FIGURE 9 |** Plotted tensor pairs for case $E_{600}$ with lowest and highest FID values.

Moreover, the trained $\mathcal{M}_1$ model is used to generate 256 $[a_{11f}]_b$ for each $E_{235}$ and $E_{600}$ case to extract into the data pool as discussed before. Then, the FID scores are computed for the generated between the real tensors. Subsequently, the Probability Density Function (PDF) is plotted to visualize the density differences for each case in **Figure 5**. It is seen that the FID scores for case $E_{600}$ are cumulated around 0.0007 with lower variance, which means that the model learned the dataset in depth and narrowed the features in the synthetic data that looks similar to the original input. For case $E_{235}$, the PDF is cumulated around 0.0025 with higher variance, which means that the model's entropy is larger; in other words, more uncertainty involves in the learned dataset and the model needs to be trained more. In summary, the generated outputs in case $E_{600}$ are very similar to the real data and can be concluded that the model learned training data.

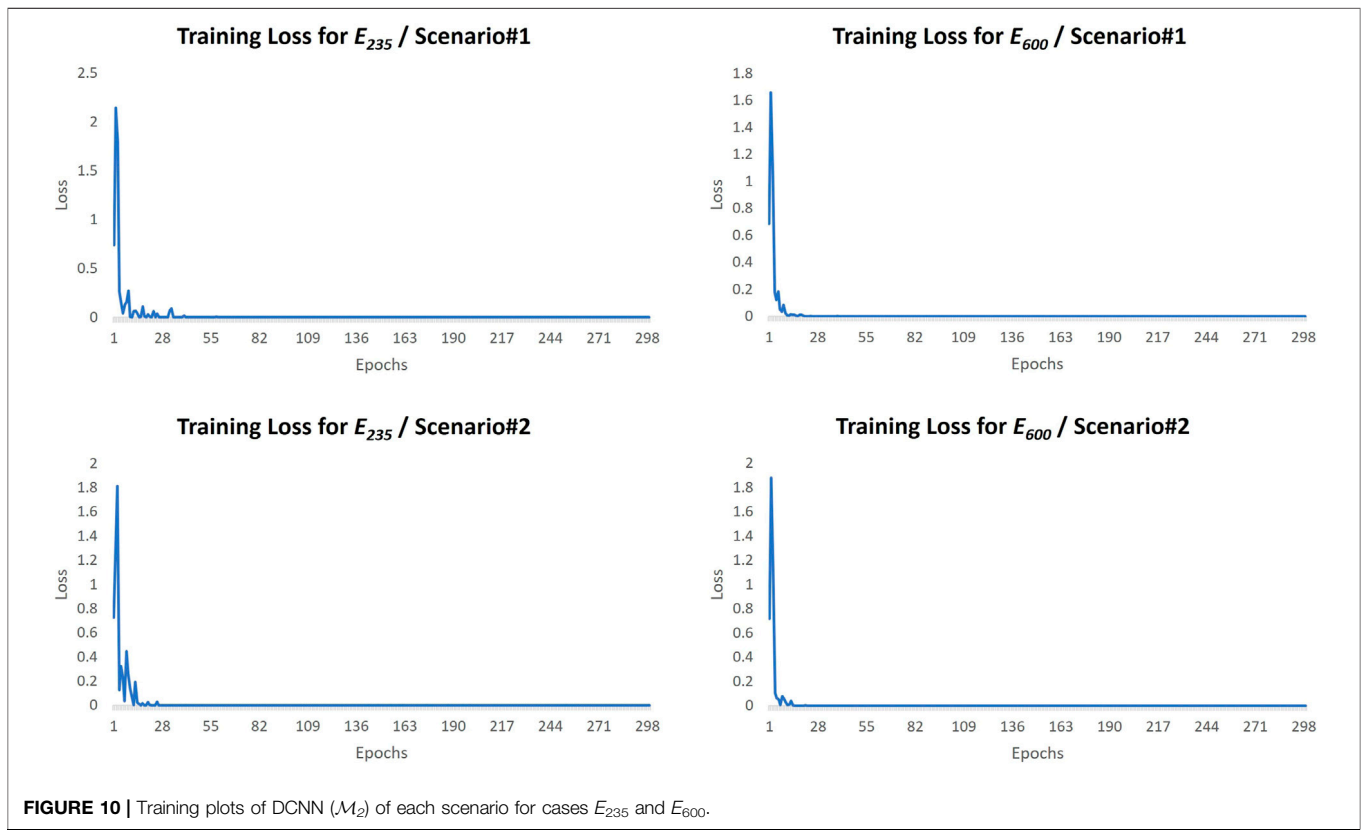

**FIGURE 10 |** Training plots of DCNN ($\mathcal{M}_2$) of each scenario for cases $E_{235}$ and $E_{600}$.

Next, the creativity and diversity measures are investigated for both cases, $E_{235}$ and $E_{600}$. First, the SSIM between the $[a_{11f}]_b$ and $[a_{11}]_b$ is computed, and the PDFs of the SSIM results are plotted in **Figure 6**. The SSIM does not go above 0.8 threshold value, which can be concluded that the generated tensors, $[a_{11f}]_b$, are not the exact copies of the real tensors, $[a_{11}]_b$; thus, $\mathcal{M}_1$ can generate creative outputs. **Figure 6** indicates that, in case $E_{600}$, the generated tensors are more similar to the real tensors as the SSIM values are cumulated around 0.3, and this value is 0.23 in case $E_{235}$. This measure also determines the overfitting of the model, which is not the case for our model here because no exact copies exist because the computed values are significantly lower than the threshold value of 0.8. Secondly, the SSIM within the generated tensors, $[a_{11f}]_b$, is computed to investigate the diversity of the generated tensor and found that the tensors in case $E_{600}$ are slightly more similar to each other than the one in case $E_{235}$. This can be interpreted as, because the $\mathcal{M}_1$ in case $E_{600}$ is trained more, it learned the data more than the $\mathcal{M}_1$ in case $E_{235}$. From the SSIM results for both cases for creativity and diversity investigation purposes, the conclusion can be made that none of the generated and real tensors are the exact copies of each other nor the generated tensors of each other.

As the last step of the quantitative evaluation of the results, the box plots are used for both cases, $E_{235}$ and $E_{600}$, separately. For that, one with the lowest and one with the highest FID scores from the created real and generated tensors are found, and the tensors are selected. Then, the found real and generated tensors,

$[a_{11}]_b$ and $[a_{11f}]_b$, respectively, that give the lowest and highest FID scores are box plotted in **Figure 7**. As the $\mathcal{M}_1$ is trained more, the variance of the generated dataset gets closer to the real dataset along with the maximum and minimum values. It can be observed that the lower the FID score is, the more box sizes and whiskers of the generated and real data are getting similar. As a result, the box plots show that the statistical meaning and distributions of the generated and real data pairs are looking very similar, especially in case $E_{600}$.

To conduct a qualitative evaluation, which is the most preferred and effective method for evaluating the image data (yet it suffers from its limitations as previously mentioned), the same vibration tensor pairs that were used in **Figure 7** are plotted in **Figures 8** and **9**. Although it is not trivial to judge the similarity between the tensors as doing it on the 2-D image data, there is a good consistency between each tensor pair.

## $\mathcal{M}_2$–Data Processing

Unlike the procedure in the $\mathcal{M}_1$—Data Processing section, the tensors from the data pool, $[a_{11}]_b$, $[a_{11f}]_b$, and $[a_{01}]_b$, are normalized between −1 and +1 range before feeding into $\mathcal{M}_2$. Subsequently, the tensors are randomly extracted without replacing from the data pool as designated order in **Figure 2** for the training of $\mathcal{M}_2$.

## $\mathcal{M}_2$–Architecture

The architecture used in $\mathcal{M}_2$ is the same architecture that is used in $\mathcal{M}_1$'s critic network. In addition, as there was no activation function used at the end of the last layer in critic in the $\mathcal{M}_1$ process (only realness or fakeness scores were used in critic), the sigmoid function is used in the $\mathcal{M}_2$ phase, which results in prediction score in a range of 0–1 where 0 is labeled as undamaged and 1 as damaged in this work. Last, no dropout is used because there is no use of it for the purpose of a simple detection procedure.

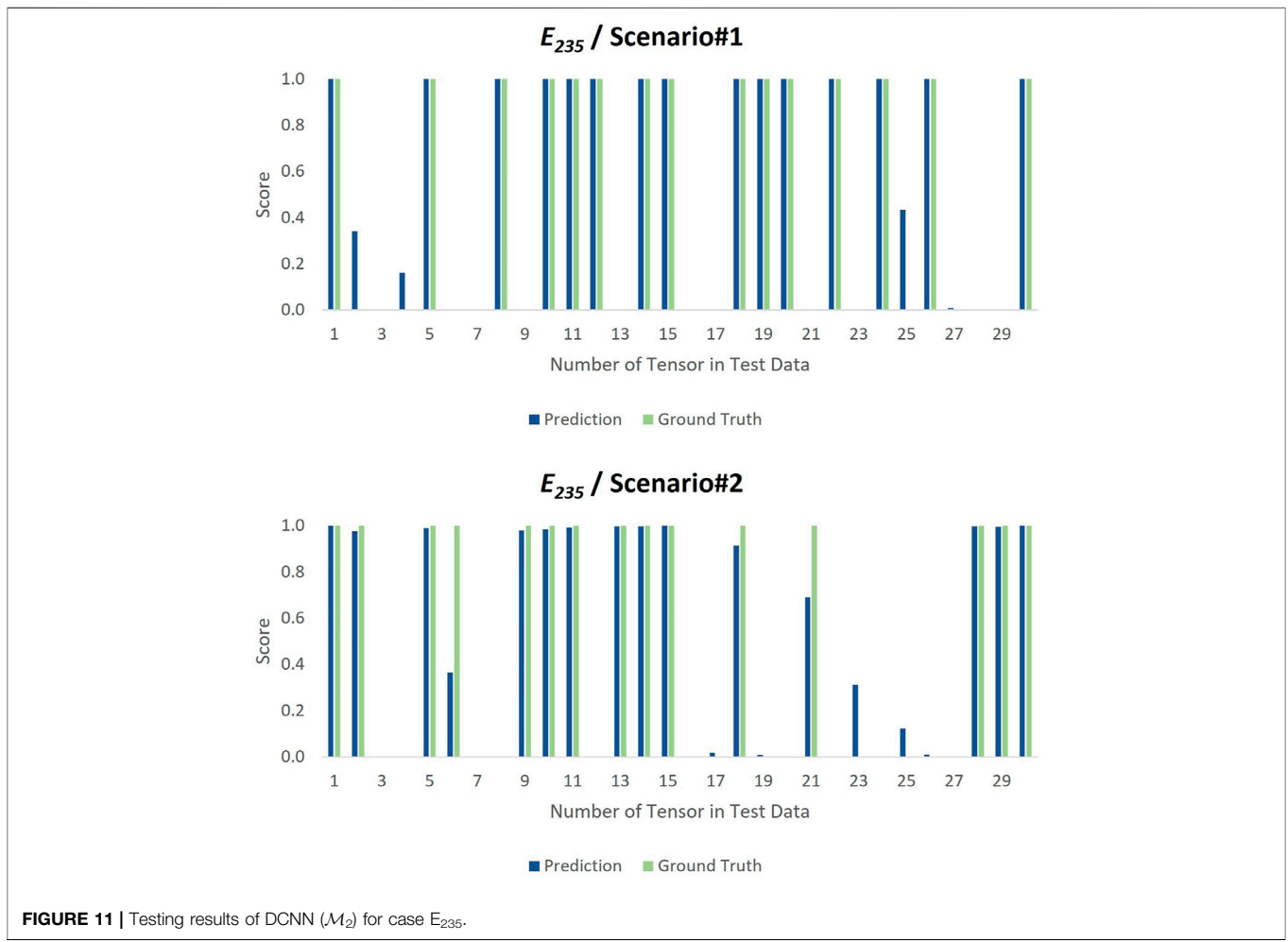

**FIGURE 11** | Testing results of DCNN ($\mathcal{M}_2$) for case E$_{235}$.

## $\mathcal{M}_2$—Training and Fine-Tuning

Like in the procedure of $\mathcal{M}_1$, the parameters are chosen for $\mathcal{M}_2$ after many trials. For cases E$_{235}$ and E$_{600}$ and Scenario#1 and Scenario#2, the used learning rate, batch size, and number of the epoch are $8 \times 10^{-4}$, 30, and 300, respectively. AdamW and Binary Cross Entropy are used as an optimizer and a loss function, respectively. Note that there was no loss function used in $\mathcal{M}_1$. Last, the training of the $\mathcal{M}_2$ model is carried out successfully as all the loss values are converged to zero for both scenarios of both cases, and the loss graphs are plotted in **Figure 10**, and the testing of the model is performed successfully on the 30 different tensors in each scenario for both cases, as designated in **Figure 2**. The evaluation of the results of training and testing the $\mathcal{M}_2$ is explained in the next section.

## $\mathcal{M}_2$—Evaluation and Interpretation of Results

It is seen in the **Figure 10** that, during the training of the $\mathcal{M}_2$ model, the training loss functions are converged to zero in each scenario for both cases; in other words, the model learned the data effectively. Yet, the critical part is in testing the model, that is, its performance on unseen data. For evaluating the results, one regression and one classification metric are used. The classification metrics such as Classification Accuracy (CA), which is the ratio of total correct predictions over total predictions, cannot be sufficient when an in-depth evaluation of the model's performance is needed. Because classification converts the prediction score into the closest label (0 or 1) based on the assumed threshold value, it might not reflect a good accuracy. For instance, a prediction score of 0.48 is converted into 0 (undamaged) if the assumed value of 0.49 as the threshold is used for classification. This might not reflect the real performance of the model because that prediction score of 0.48 might be a result of 48% of damage (as a quantification of the damage such as 48% of bolt loosened) in the system or might be a lot higher or lower and classifying it as label 0 can be a faulty prediction. A ROC and AUC curve can be employed to look for the optimum threshold value that gives the desired classification results for damage detection. Yet, this is not in the scope of this study, and there are not many classification samples in the testing data; therefore, it is not necessary to check for all different thresholds. Thus, the threshold in this study is defined as 0.49 for the CA. Moreover, it is possible for DL-based structural damage diagnostics that the probability of the tensor being undamaged or damaged can also mean the quantification result of the tensor whether it is undamaged or damaged. This is another area of investigation for different levels (detection, localization, and quantification) of damage diagnostics using DL models. Along with a CA metric, a regression metric, Mean Absolute Error (MAE). MAE measures the average of all the prediction errors by taking the summation of the absolute values of the predicted values minus the actual values and then dividing it by the total sample in the test. MAE is an excellent metric tool to measure the model's error on the dataset.

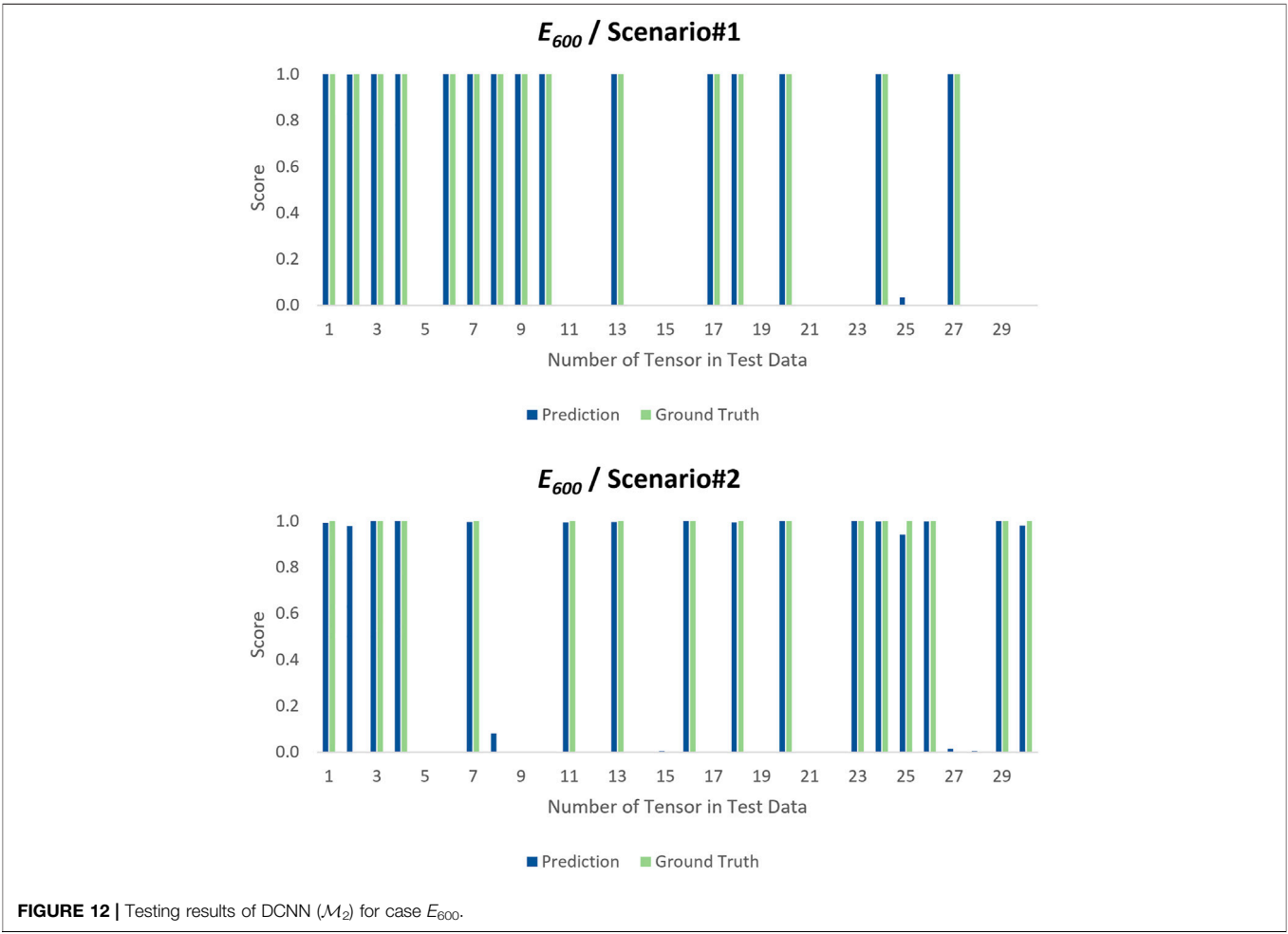

FIGURE 12 | Testing results of DCNN ($\mathcal{M}_2$) for case $E_{600}$.

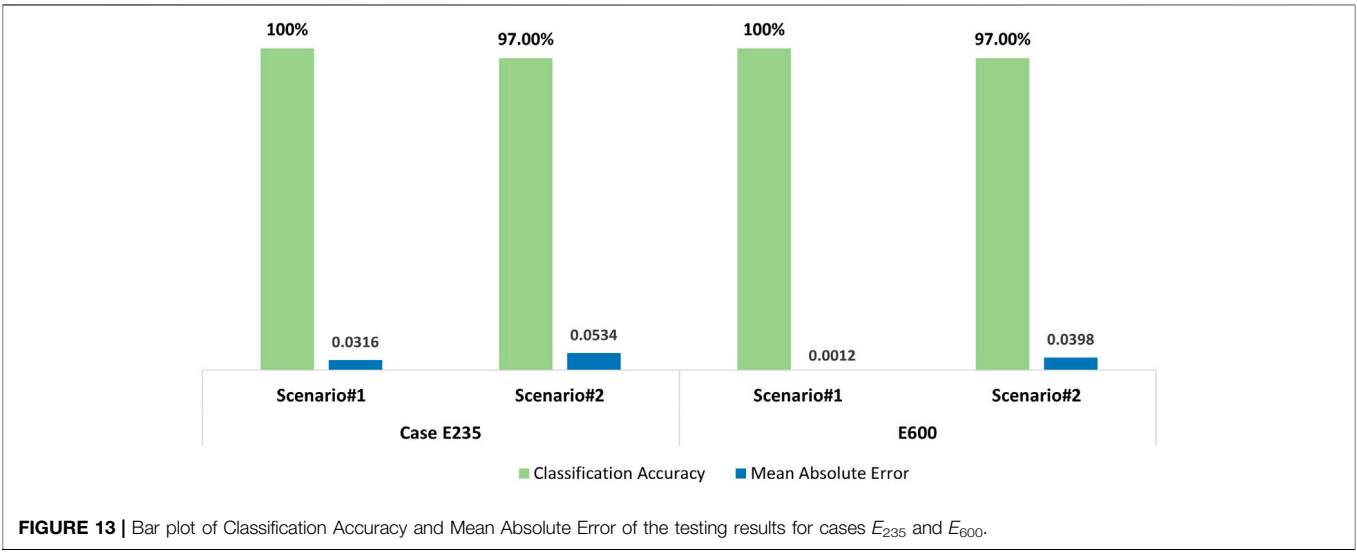

FIGURE 13 | Bar plot of Classification Accuracy and Mean Absolute Error of the testing results for cases $E_{235}$ and $E_{600}$.

The resulted prediction scores along with the ground truths (correct target value) on the test tensors are bar plotted in **Figures 11** and **12** for cases $E_{235}$ and $E_{600}$, including each of their scenarios, respectively. At first glance, it is easy to catch the close prediction scores to ground truths, particularly for case $E_{600}$ where the scores are closer because the $\mathcal{M}_2$ is trained more. Likewise, the prediction scores are slightly closer to the ground truths in Scenario#1 in both cases because no synthetic tensor is involved in the process. Yet, more training helped the model to learn more about the data; thus, it could identify the test instances with more accuracy as this is the situation in both scenarios in case $E_{600}$. In **Figure 11**, for Scenario#1, $\mathcal{M}_2$ is looking very consistent and confident on the test instances as it predicted all of them successfully. For case $E_{600}$, Scenario#2, where the model is tested on only synthetic damaged and real undamaged instances, has a slightly off prediction score around 0.10 on tensor 8, which might be negligible. Yet, the quantitative results can give better insight into the model's performance. Hence, the results are reflected on the CA and MAE metrics, as in **Figure 13**, and it is seen that the CAs are following the same scores, 100% and 97%, for both cases and scenarios, respectively. Considering that the CA results are very high with only one incorrect prediction, which is for undamaged tensor as damaged, the resulting values in this paper can be concluded as excellent. Yet, the criterion to consider a success rate as "excellent" mainly depends on the used application of the model. For example, in damage detection problems, in order for a classification value of the model's performance to be considered acceptable, it is prominent to know what percentage of the dataset can be ignored for an achievable safety measure. Last, the MAE results are seemingly very low, which implies that $\mathcal{M}_2$ has very low error values on the unseen data, and the model can predict the damage-associated instances from the undamaged data successfully.

## CONCLUSION

Detecting and locating damage on large civil structures is a challenging task and has been the subject of research for many years. Finding damage-associated vibration data from structures can also be very challenging. Therefore, data scarcity is a setback in civil SHM applications. With the rapid developments in AI, ML, and DL, researchers started to take advantage of such tools for vibration-based structural damage diagnostics and achieved great success. Considering that the DL algorithms perform better with large data, this study used 1-D WGAN-GP to generate synthetic vibration data and validate the produced outputs by employing 1-D DCNN trained on real data and tested on a synthetic dataset. The same DCNN model is also tested on a real dataset for benchmarking purposes. The classification results showed that the performance of DCNN on the test data and on the real data is 97% and 100%, respectively, with both scenarios resulting in significantly low model errors. The main conclusions of this study can be listed in bullet points:

- The well-known problem of data scarcity in SHM of civil structures can be tackled by using GANs to generate similar types of datasets. This study showed that GANs could produce a vibration signal that is almost indistinguishable from the real ones, and the generated signals are verified with metrics. Although this study used one of the most advanced GAN methods, the WGAN-GP algorithm, which improves the well-known training problems of GANs, the training is still tedious.

- The study demonstrated that the generated vibration signals are indistinguishable by DCNN. This means that, in a situation where some damage-associated dataset exists for a civil structure, yet the amount of existing dataset is not sufficient for vibration-based damage diagnostics via DL model, and then, the proposed methodology can be used as a solution. As a result, the presented methodology paves the way for more ML- and DL-based models to be utilized for structural damage diagnostics.

- The generated data from GAN is creative and versatile enough that it is not a copy of the input nor a copy of other generated datasets. The analogy is indeed similar to the real structures where dynamic responses for different damage types can contain similar characteristics in the raw data, yet they are not copies of each other. Therefore, when the data are scarce, or a portion of it is missing, GANs can generate additional data similar "enough" to the original data. It also diversifies the input data by adding a somewhat new meaning to the learned variation range of the dataset, which can be used for various damage detection problems for multiple civil structures by adjusting the features in the data used. Hence, it can be used successfully for vibration-based nonparametric damage diagnostics, as demonstrated in this study.

- The presented methodology's use-case scenario is utilizing a DL model, e.g., 1-D DCNN on a multi-span bridge where only one span has a damaged-associated dataset. To implement damage diagnostics on the raw vibration data using a DL model, the model has to be trained with the already existing data samples, which are undamaged and damaged-associated datasets (undamaged and damaged classes). Yet, it is important to note that the damaged-associated datasets are comparatively very few. This class imbalance in the training dataset of the DL model lowers the model's performance. In other words, to tackle this data scarcity problem, WGAN-GP can be used to generate the needed data samples to train the DL model efficiently. Thus, the methodology increases the performance of the DL model.

- On the basis of the demonstrated performance herein by GANs on generating vibration data, there is good potential for further use of them in civil SHM applications. More research is needed to generate the damaged dynamic response of structures using the undamaged responses.

## DATA AVAILABILITY STATEMENT

The datasets presented in this study can be found in online repositories. The names of the repository/repositories and accession number(s) can be found below: http://www. structuraldamagedetection.com/benchmark/.

## AUTHOR CONTRIBUTIONS

FC developed the proposal and obtained research funding for the study. FC, OA, and FL contributed to conception and design of the study. FL organized the database and performed the data analysis. FC and OA developed the paper outline, and FL wrote the first draft of the manuscript. All authors contributed to manuscript revision, read, and approved the submitted version.

## ACKNOWLEDGMENTS

The authors would like to thank members of CITRS (Civil Infrastructure Technologies for Resilience and Safety) Research Initiative at the University of Central Florida. The second author would like to acknowledge the support by National Aeronautics and Space Administration (NASA) Award No. 80NSSC20K0326.

**Conflict of Interest:** The authors declare that the research was conducted in the absence of any commercial or financial relationships that could be construed as a potential conflict of interest.

# NOMENCLATURE

$[a_{01}]_r$ Vibration data at joint#1 in undamaged scenario (0) in raw form (262,144 samples)

$[a_{11}]_r$ Vibration data at joint#1 in damaged scenario (1) in raw form (262,144 samples)

$[a_{11}]_b$ Vibration tensor at joint#1 in damaged scenario (1) in batched form (1,024 samples)

$[a_{01}]_b$ Vibration tensor at joint#1 in undamaged scenario (0) in batched form (1,024 samples)

$[a_{11f}]_b$ Generated, "fake", vibration tensor from GAN which the model is trained on $[a_{11}]_r$

$E_{235}$ Case where the GAN trained for 235 epochs

$E_{600}$ Case where the GAN trained for 600 epochs

$\mathcal{M}_1$ Used 1-D WDCGAN-GP model in the paper

$\mathcal{M}_2$ Used 1-D DCNN model in the paper